# How to Train Your Deep Neural Network with Dictionary Learning

Vanika Singhal[*], Shikha Singh[+] and Angshul Majumdar[#]

[*]*IIIT Delhi*
*Okhla Phase 3*
*Delhi, 110020, India*
vanikas@iiitd.ac.in

[+]*IIIT Delhi*
*Okhla Phase 3*
*Delhi, 110020, India*
shikhas@iiitd.ac.in

[#]*IIIT Delhi*
*Okhla Phase 3*
*Delhi, 110020, India*
angshul@iiitd.ac.in

*Abstract*: Currently there are two predominant ways to train deep neural networks. The first one uses restricted Boltzmann machine (RBM) and the second one autoencoders. RBMs are stacked in layers to form deep belief network (DBN); the final representation layer is attached to the target to complete the deep neural network. Autoencoders are nested one inside the other to form stacked autoencoders; once the stcaked autoencoder is learnt the decoder portion is detached and the target attached to the deepest layer of the encoder to form the deep neural network. This work proposes a new approach to train deep neural networks using dictionary learning as the basic building block; the idea is to use the features from the shallower layer as inputs for training the next deeper layer. One can use any type of dictionary learning (unsupervised, supervised, discriminative etc.) as basic units till the pre-final layer. In the final layer one needs to use the label consistent dictionary learning formulation for classification. We compare our proposed framework with existing state-of-the-art deep learning techniques on benchmark problems; we are always within the top 10 results. In actual problems of age and gender classification, we are better than the best known techniques.

## 1. Introduction

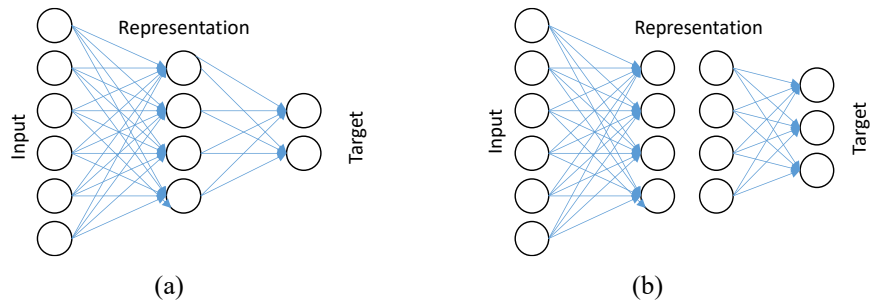

**Figure 1.** a. Single Representation Layer Neural Network. b. Segregated Representation

The schematic diagram (Figure 1a) shows a shallow neural network with a single hidden layer. Such a neural network is trained with training samples at the input and the corresponding (usually binarized) class labels at the output targets. It learns the network weights by backpropagating the error.

Training the shallow neural network can be perceived as a segregated problem – learning weights between the input and the hidden / representation layer and between representation layer and the output / targets. If the network between the input and the representation layer is already learnt, training the second network is trivial. It is a simple regression problem since both the input (representation of training samples) and the outputs are known. Training the first layer of network weights between the input and the representation is the challenging task, since two variables (the network weights and the representation) need to be learnt from the input training samples. This (training the first layer) is the topic of 'representation learning'.

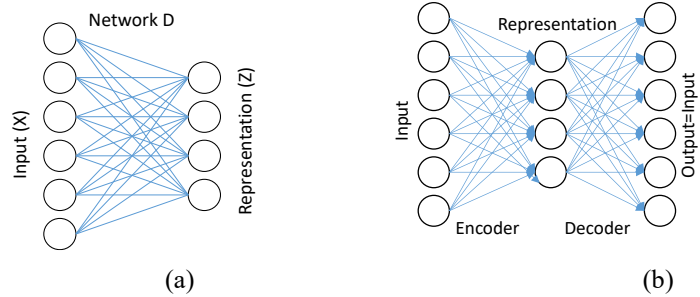

Figure 2. a. Restricted Boltzmann Machine. b. Autoencoder

The main concern while training the representation layer is that the information content of the input must be preserved. Currently there are two popular ways to train the representation layer. One such approach is via Restricted Boltzmann Machine (RBM) [1]. RBM is an undirected graphical model that uses stochastic hidden units to model the distribution over the stochastic visible units. The hidden layer is symmetrically connected with the visible unit, and the architecture is "restricted" as there are no connections between units of the same layer. Traditionally, RBMs are used to model the distribution of the input data *p(x)*.

The schematic diagram of RBM is shown in Figure 2a. The objective is to learn the network weights (*W*) and the representation (*H*). This is achieved by optimizing the Boltzmann cost function given by:

$$p(W,H) = e^{-E(W,H)} \tag{1}$$

where, $E(W,H) = -H^T W X$ including the bias terms.

Broadly speaking, RBM learning is based upon maximizing the similarity between the projection of the data and the representation, subject to the usual constraints of probability. In RBM, the information content of the input is preserved in the sense of maximizing similarity. Once the RBM is learnt, it is used as the first part of a single layer neural network. Once the targets are attached to the output of the RBM (and network weights learnt) it forms a complete neural network.

The second popular technique for representation learning is the autoencoder [2] (Figure 2b). It consists of two parts – the encoder maps the input to a latent representation and the decoder maps the latent representation back to the data. For a given input vector (including the bias term) *x*, the latent space is expressed as:

$$h = \phi(Wx) \tag{2}$$

Here $\phi$ is the non-linear activation function. The decoder reverse maps the representation to the data space – hence the name 'autoencoder' or 'auto associative memory'.

$$x = W'\phi(Wx) \tag{3}$$

Since the data space is assumed to be the space of real numbers, there is no sigmoid function here. During training, the problem is to learn the encoding and decoding weights – W and W'. These are learnt by minimizing the Euclidean cost:

$$\arg\min_{W,W'} \|X - W'\phi(WX)\|_F^2 \qquad (4)$$

In autoencoder, the information is preserved at the representation in the Euclidean sense, such that the inputs can be recovered with minimal $l_2$-norm loss.

For forming a neural network, the decoder portion of the autoencoder is removed. The encoder acts as the first layer (input to representation) of the neural network (Figure 1b). The targets are attached to the representation and the corresponding weights are learnt to complete the neural network training.

In recent times the Extreme Learning Machine (ELM) [3] is also gaining popularity. It is a single layer neural network where the network weights between the input and the representation layer are randomly assigned values. Therefore there is no representation learning required. The second layer between the representation and the output is learnt in closed form by minimizing the Euclidean loss. ELM is not the topic of discussion; but we mention it for the sake of completeness.

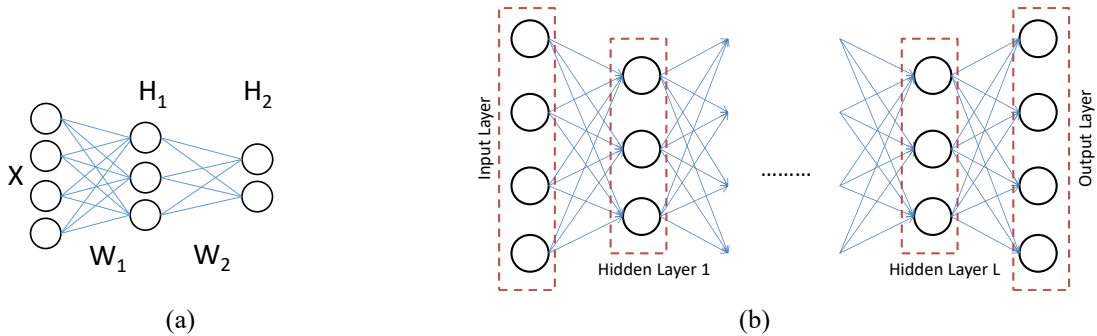

**Figure 3.** a. Deep Boltzmann Machine. b. Stacked Autoencoder

Usually training a single layer neural network is easy; one does not employ RBM and autoencoder for training such shallow neural networks. Deep neural networks have multiple representation layers. In such a case training them directly becomes difficult in practice. Representation learning techniques are used in such cases. One can either build a deep neural network using RBM as the basic units or autoencoders.

RBMs can be stacked one after the other to form deep Botlzmann machine (DBM) (Figure 3a) [4]. DBM is undirected. There can be a directed model arising from stacking RBM leading to deep belief network (DBN) [5]; this is more attuned towards neural networks. The targets are attached to the final layer of the DBN and the weights between the final representation layer and he target is learnt – thereby completing the training of the deep neural network.

Deep networks can also be formed by stacking one autoencoder inside the other. This is shown in Figure 2b. These are called stacked autoencoders [5]; they have multiple levels of encoders and the same number of decoders. Once the stacked autoencoder is learnt, the decoder portion is detached and the targets attached to the representation of the deepest layer. This forms the deep neural network (once the weights between the deepest representation layer and the target is learnt).

There are convolutional neural network (CNN) based deep learning models as well. They yield amazing results, but they are restricted mostly to imaging problems. Our interest lies in generic deep neural networks and hence CNN will not be discussed.

In this work we will show how dictionary learning can be used as a representation learning tool and deep neural networks be built with dictionary learning as basic units. The proposed framework will be pitted against the best deep learning architectures on benchmark problems; we will see how our simple framework features among the top 10 methods. The framework has also been applied to the problem of face image based age and gender classification; we yield better results than the best known techniques.

## 2. Proposed Dictionary Based Deep Neural Network

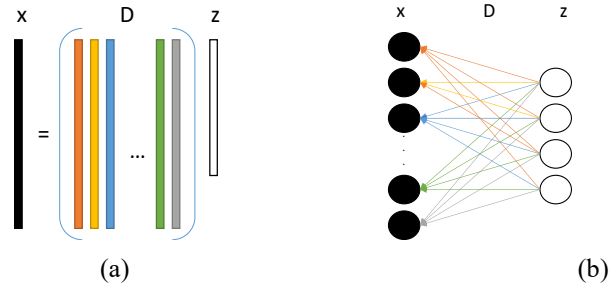

(a)                                            (b)

**Figure 4.** a. Dictionary Learning. b. Our Neural Network Interpretation

The usual interpretation for dictionary learning is different is that it learns a basis ($D$) for representing ($Z$) the data ($X$) (Figure 4a). The columns of $D$ are called 'atoms'. In this work, we look at dictionary learning in a different manner. Instead of interpreting the columns as atoms, we can think of them as connections between the input and the representation layer. To showcase the similarity, we have kept the color scheme intact in Figure 4b.

Unlike a neural network which is directed from the input to the representation, the dictionary learning kind of network points in the other direction – from representation to the input. Dictionary learning employs an Euclidean cost function (2), given by

$$\min_{D,Z} \|X - DZ\|_F^2 \tag{5}$$

This is easily solved using alternating minimization of the dictionary $D$ and the codes $Z$. Today most studies (following K-SVD [7]) impose an additional sparsity constraint on the codes ($Z$), but it is not mandatory.

Note that dictionary learning indeed follow the basic premise of representation learning. The information content of the inputs ($X$) are preserved in the features $Z$ in the Euclidean sense.

Based on the neural network type interpretation of dictionary learning there are a handful of prior studies that proposed techniques to learn deeper features [8, 9]. The first layer of dictionary learns from the input training data. The subsequent layers learn from the features from the previous layer as inputs. The prior studies only proposed a representation learning tool. They did not learn a complete neural network. In this work, we show how a deep neural network can be learnt with a plug-and-play approach.

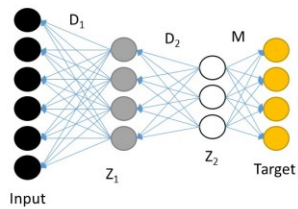

**Figure 5.** Deep Neural Network with Dictionary Learning.

The deep neural network is shown in Figure 5. Since dictionary learning is a synthesis approach the arrows are pointed (in the opposite direction) from the representation to the input for the representation layers ($Z_1$ and $Z_2$). But for the final layer – between the final level of representation to the target the arrows point in the usual direction. For such a network we write the cost function as:

$$\min_{D_1,...D_N,Z,M} \|X - D_1\varphi(D_2\varphi(...\varphi(D_N Z)))\|_F^2 + \|T - MZ\|_F^2 \qquad (6)$$

Here $D_1$ to $D_N$ are the $N$ level dictionaries, $Z$ the final level representation, $T$ the targets and M the linear map from the representation to the targets.

Solving (6) exactly is a difficult problem. The difficulty arises in training all deep neural networks. To circumvent this, a greedy approach layer-by-layer training approach is followed [10]. We follow a greedy approach as well.

For the first layer of dictionary learning, we can express $Z_1 = \varphi(D_2\varphi(...\varphi(D_N Z)))$. Therefore greedy learning of the first layer is represented by,

$$\min_{D_1,Z_1} \|X - D_1 Z_1\|_F^2 \qquad (7)$$

This is a typical dictionary learning formulation. For the second layer we have $Z_1 = \varphi(D_2\varphi(...\varphi(D_N Z)))$; we substitute $Z_2 = \varphi(D_3...\varphi(D_N Z))$. This allows the expression

$$\varphi^{-1}(Z_1) = D_2 Z_2 \qquad (8)$$

It is easy to invert the activation function since it operates element-wise. This allows solving (8) via dictionary learning.

$$\min_{D_2,Z_2} \|\varphi^{-1}(Z_1) - D_2 Z_2\|_F^2 \qquad (9)$$

One may argue about values in $Z$ that would make the output of $\varphi^{-1}$ to be infinity. The problem arises in any neural network. Recently an elegant solution has been proposed in [11] – that of adding slight amount of noise; we follow the same here.

With such substitutions and dictionary learning we can learn until the penultimate layer. In the final layer we will have for the representation learning term: $Z_{N-1} = \varphi(D_N Z) \Rightarrow \varphi^{-1}(Z_{N-1}) = D_N Z$. This would lead to a cost function of the form

$$\min_{D_N,Z} \|\varphi^{-1}(Z_{N-1}) - D_N Z\|_F^2 \qquad (10)$$

But there is also the term for mapping the representation to the targets – the second term is (6). Therefore, we need to add it to (10). The final level of joint representation and linear map learning is therefore expressed as,

$$\min_{D_N, Z, M} \|\varphi^{-1}(Z_{N-1}) - D_N Z\|_F^2 + \|T - MZ\|_F^2 \tag{11}$$

Although not a standard dictionary learning formulation (11) is a solved problem. It is known as label-consistent KSVD [12] in computer vision literature.

We have shown how the complex problem (6) can be segregated into smaller sub-problems that have well-defined solutions in dictionary learning literature. For all the layers till the final one a simple alternating minimization algorithms such as method of optimal directions [13] of multiplicative updates [14] can be used. For the last layer we use LC-KSVD [12].

There are several advantages of layer-wise learning:

1. For each layer, well-tested dictionary learning algorithms are available.

2. Learning the deep network is one go, requires solving a multitude of parameters. With limited training data, learning so many networks leads to over-fitting. For greedy layer-wise learning the number of parameters to learn in each stage is relatively small. So the issue of over-fitting is less pronounced.

3. There are certain mathematical guarantees for shallow dictionary learning [15]. These guarantees will be hard to generalize for multiple layers.

## 2.1. Plug-and-play Approach

So far we have discussed the use of standard dictionary learning for each of the layers. This leads to the basic deep neural network architecture, but there is scope of further improvement. In all the layers till the final layer the dictionary learning is unsupervised. Since each of the layers are learnt separately we can follow a plug-and-play approach for learning these layers. We can pick up any supervised dictionary learning technique and use it to generate features at each of the levels (before the final layer).

A few such examples of supervised dictionary learning will be given here. However there is a plethora of literature on this topic and given the limitations of space we cannot be encyclopedic in coverage.

One of the first studies in supervised learning was proposed in [16]. Here on top of the dictionary learning cost function there is an extra term that accounts for classification error:

$$\min_{D, Z, \theta} \|X - DZ\|_F^2 + \lambda C(y, f(Z, \theta)) \tag{12}$$

where $y \in \{-1, 1\}$ and $C(x) = \log(1 + e^{-x})$ is the classification error penalty that is very similar to the hinge loss used in SVM; $f(\theta, Z) = \theta Z + b$.

When used in our plug-and-play deep learning framework, this technique is especially suitable for solving binary classification problems. The features generated at level will be optimally separated for two classes. There are many other formulations for binary classification using dictionary learning; for example [17] uses a Fisher linear discriminant criterion. Owing to limitations in space we cannot discuss all such methods.

In [18] a technique is proposed to address the multi-class feature learning problem. It learns a separate dictionary for each class. The training samples are expressed as,

$$X = D_1 Z_1 + \ldots + D_C Z_C + D_S Z_S \tag{13}$$

C classes are assumed here. $D_1$ to $D_C$ are the class specific dictionaries, $D_S$ is the shared dictionary by all classes. $Z_1$ to $Z_C$ are the features for each class and $Z_S$ the shared representation. To make the representation discriminative [18] enforced that $Z_{i^c} = 0$; It means that the non-zero coefficients of samples $X_i$ will only concentrate on the sub-dictionaries $D_i$ and $D_s$, while the class-specific sub-dictionary $D_i$ will be having explicit correspondence to class labels i. The learning is formulated as,

$$\min_{D,Z} \|X - DZ\|_F^2 + \sum_i \|X_i - D_i Z_i - D_S Z_{i,s}\|_F^2 + \eta \sum_{i \neq j} \|D_i^T D_j\|_F^2 \tag{14}$$

The first two terms are the discriminative fidelity term. The last term is the mutual incoherence term between the dictionaries of every class. An improvement of this techniques was proposed in [19].

There are several other formulations for multi-class supervised dictionary learning. It is not possible to discuss all of them. However, we can pick up any suitable formulation and instead of the unsupervised formulation in the pre-final layers, we can plug a multi-class supervised dictionary learning techniques.

For the final layer we can use the simple LC-KSVD formulation as discussed before, or we can use a slightly advanced version of it (dubbed LC-KSVD2 [12]) where class specific atoms are learnt in the dictionary. This is given by,

$$\min_{D_N, Z, M, W} \|\varphi^{-1}(Z_{N-1}) - D_N Z\|_F^2 + \mu \left( \|T - MZ\|_F^2 + \|H - WZ\|_F^2 \right) \tag{15}$$

H is a 'discriminative' sparse code corresponding to an input signal sample, if the nonzero values of $H_i$ occur at those indices where the training sample $X_i$ and the dictionary item $D_{(N)k}$ share the same label.

## 3. Experimental Results

### 3.1. Experiments on Benchmark Deep Learning Datasets

In this work we report results on object recognition benchmarking datasets – MNIST (error), CIFAR-10 (accuracy), CIFAR-100 (accuracy) and SVHN (error). We only compare with prior published works (not including manuscripts in arxiv). Since all of them are multi-class problems we try two variants of our proposed dictionary based deep neural network (DDNN). In the first one (DDNN1) unsupervised dictionary learning is used till the pre-final level; the final level uses LC-KSVD1. In the second variant (DDNN2) discriminative dictionary learning from [19] is used till the pre-final level; the final uses LC-KSVD2. Both variants use a three layer architecture; the number of dictionary atoms are halved in every layer. For the second variant, the number of atoms assigned to each dictionary is each layer is uniformly distributed across the classes.

Since the said datasets have defined protocols, we just compare it with the results assembled by Rodrigo Beneson [20]; the results are shown in Tables 1-4. We find that our proposed techniques are always within the top 10. Most of the techniques in the following tables are based on CNN – it requires significant hand tuning and heuristic parameter optimization. Our method is simple and straightforward and yet we perform at par or better than the most. We believe that using convolutional dictionary learning layers in the initial stages can boost the results even further.

Table 1. MNIST

| Result | Method | Venue |
|---|---|---|
| 0.21% | Regularization of Neural Networks using DropConnect | ICML 2013 |
| 0.23% | Multi-column Deep Neural Networks for Image Classification | CVPR 2012 |
| 0.29% | Generalizing Pooling Functions in Convolutional Neural Networks: Mixed, Gated, and Tree | AISTATS 2016 |
| **0.31%** | **DDNN2 (Proposed)** | |
| 0.31% | Recurrent Convolutional Neural Network for Object Recognition | CVPR 2015 |
| 0.35% | Deep Big Simple Neural Nets Excel on Handwritten Digit Recognition | Neural Computation 2010 |
| 0.39% | Efficient Learning of Sparse Representations with an Energy-Based Model | NIPS 2006 |
| **0.40%** | **DDNN1 (Proposed)** | |
| 0.40% | Best Practices for Convolutional Neural Networks Applied to Visual Document Analysis | DAR 2003 |

Table 2. CIFAR-10

| Result | Method | Venue |
|---|---|---|
| 95.59% | Striving for Simplicity: The All Convolutional Net | ICLR 2015 |
| 94.16% | All you need is a good init | ICLR 2015 |
| 93.95% | Generalizing Pooling Functions in Convolutional Neural Networks: Mixed, Gated, and Tree | AISTATS 2016 |
| 93.63% | Scalable Bayesian Optimization Using Deep Neural Networks | ICML 2015 |
| **93.08%** | **DDNN2 (proposed)** | |
| 92.91% | Recurrent Convolutional Neural Network for Object Recognition | CVPR 2015 |
| 92.51% | Learning Activation Functions to Improve Deep Neural Networks | ICLR 2015 |
| 92.40% | Training Very Deep Networks | NIPS 2015 |
| 91.88% | Multi-Loss Regularized Deep Neural Network | CSVT 2015 |
| **91.77%** | **DDNN1 (Proposed)** | |

Table 3. CIFAR-100

| Result | Method | Venue |
|---|---|---|
| 72.60% | Scalable Bayesian Optimization Using Deep Neural Networks | ICML 2015 |
| 72.34% | All you need is a good init | ICLR 2015 |
| 69.17% | Learning Activation Functions to Improve Deep Neural Networks | ICLR 2015 |
| **68.82%** | **DDNN2 (Proposed)** | |
| 68.53% | Multi-Loss Regularized Deep Neural Network | CSVT 2015 |
| 68.40% | Spectral Representations for Convolutional Neural Networks | NIPS 2015 |
| 68.25% | Recurrent Convolutional Neural Network for Object Recognition | CVPR 2015 |
| **68.00%** | **DDNN1 (Proposed)** | |
| 67.76% | Training Very Deep Networks | NIPS 2015 |
| 67.68% | Deep Convolutional Neural Networks as Generic Feature Extractors | IJCNN 2015 |

Table 4. SVHN

| Result | Method | Venue |
|---|---|---|
| 1.69% | Generalizing Pooling Functions in Convolutional Neural Networks: Mixed, Gated, and Tree | AISTATS 2016 |
| 1.77% | Recurrent Convolutional Neural Network for Object Recognition | CVPR 2015 |
| **1.80%** | **DDNN2 (proposed)** | |
| 1.92% | Recurrent Convolutional Neural Network for Object Recognition | CVPR 2015 |
| 1.94% | Regularization of Neural Networks using DropConnect | ICML 2013 |
| 2.15% | BinaryConnect: Training Deep Neural Networks with binary weights during propagations | NIPS 2015 |
| **2.26%** | **DDNN1 (Proposed)** | |
| 2.35% | Network in Network | ICLR 2014 |
| 2.47% | Maxout Networks | ICML 2013 |
| 4.90% | Convolutional neural networks applied to house numbers digit classification | ICPR 2012 |

### 3.2. Experiments on Age and Gender Classification from Face Image

Adience is the benchmark dataset [21] for age and gender classification. The dataset consists of images automatically uploaded to Flickr from smart-phone devices. Because these images were uploaded without prior manual filtering, as is typically the case on media web-pages or social websites, viewing conditions in these images are highly unconstrained, reflecting many of the real-world challenges of faces appearing in Internet images. Adience images therefore capture extreme variations in head pose, lightning conditions quality, and more. The entire Adience collection includes roughly 26K images of 2,284 subjects. Testing for age and gender classification is performed using a standard five-fold, subject-exclusive cross-validation protocol, defined in [21]. We use the in-plane aligned version of the faces used there in.

We have compared our method with the very best available methods – DEX (Deep Expectation) [22] (winner of ChaLearn LAP Challenge at ICCV 2015 for age estimation) and Levi and Hassner [23] (best results on Adience). It is shown in the following table.

For our proposed formulation, we have used the DDNN1 as the base model. Since age prediction is a multi-class problem we use DDNN2. For gender prediction (being a binary classification problem) we use the FLD dictionary learning formulation [17] in each of the pre-final stages; this is the DDNN3 formulation. The number of dictionary atoms are halved in every layer.

Table 5. Age and Gender Classification Results

| Method | Age Prediction | Gender Prediction |
|---|---|---|
| Levi and Hassner [23] (over-sampling) | 50.7 | 86.8 |
| Levi and Hassner [23] (single crop) | 49.5 | 85.9 |
| DEX [22] | 46.6 | Cannot predict gender |
| **DDNN1** | **49.8** | **86.2** |
| **DDNN2 & DDNN3** | **50.5 (DDNN2)** | **87.0 (DDNN3)** |

We find that DDNN1 yields better results than the DEX [22] method for age prediction. It also uses better results than [23] when the full image is used. But [23] proposed a second formulation where patches are taken from the image; DDNN1 cannot beat this method. However our proposed supervised formulations DDNN2 and DDNN3 yields even better results than the patch based formulation proposed in [23].

## 4. Conclusion

This work proposed a new method to train deep neural networks. Prior studies used RBM or autoencoder as the basic building blocks. This work shows how dictionary learning can be used as building blocks for deep neural networks. The framework is flexible and one can build the deep network in a plug-and-play fashion. One can pick and choose any dictionary learning variant of choice for each layer. There is a plethora of dictionary learning techniques to choose from, and one has the liberty to mix and match these techniques in our proposed plug-and-play framework.

This work applies the proposed framework for training deep neural network to some computer vision problems. We show that our technique always ranks among the top few on benchmark deep learning datasets. When applied to the problem of face image based gender and age classification, we beat the state-of-the-art.

## References

[1] R. Salakhutdinov, A. Mnih and G. Hinton, "Restricted Boltzmann machines for collaborative filtering", ACM ICML, pp. 791-798, 2007.

[2] P. Baldi, "Autoencoders, unsupervised learning, and deep architectures". ICML workshop on unsupervised and transfer learning, pp. 37-50, 2012.

[3] G. B. Huang, Q. Y. Zhu, and C. K. Siew, "Extreme learning machine: theory and applications", Neurocomputing, Vol. 70 (1), 489-501, 2006.

[4] R. Salakhutdinov and G. E. Hinton, "Deep Boltzmann Machines", AISTATS, 2009.

[5] N. Le Roux and Y. Bengio, "Representational power of restricted Boltzmann machines and deep belief networks", Neural computation, Vol. 20 (6), pp. 1631-1649, 2008.

[6] P. Vincent, H. Larochelle, I. Lajoie, Y. Bengio and P. A. Manzagol, " Stacked denoising autoencoders: Learning useful representations in a deep network with a local denoising criterion". Journal of Machine Learning Research, Vol. 11, pp. 3371-3408, 2010.

[7] R. Rubinstein, A. M. Bruckstein and M. Elad, "Dictionaries for Sparse Representation Modeling", Proceedings of the IEEE, Vol. 98 (6), pp. 1045-1057, 2010.

[8] V. Singhal, A. Gogna and A. Majumdar, "Deep Dictionary Learning vs Deep Belief Network vs Stacked Autoencoder: An Empirical Analysis", ICONIP 2016.

[9] S. Tariyal, A. Majumdar, R. Singh and M. Vatsa. "Greedy Deep Dictionary Learning", arXiv:1602.00203.

[10] Y. Bengio, P. Lamblin, P. Popovici and H. Larochelle, "Greedy Layer-Wise Training of Deep Networks", NIPS, 2007.

[11] Ç. Gülçehre, M. Moczulski, M. Denil and Y. Bengio, "Noisy Activation Functions", ICML, 2016.

[12] Z. Jiang, Z. Lin and L. S. Davis, "Learning A Discriminative Dictionary for Sparse Coding via Label Consistent K-SVD", IEEE Transactions on Pattern Analysis and Machine Intelligence, Vol. 35, pp. 2651-2664, 2013.

[13] K. Engan, S. Aase, and J. Hakon-Husoy, "Method of optimal directions for frame design," IEEE ICASSP, 1999.

[14] C. J. Lin, "On the convergence of multiplicative update algorithms for nonnegative matrix factorization", IEEE Transactions on Neural Networks, Vol. 18(6), 1589-1596, 2007.

[15] S. Arora, A. Bhaskara, R. Ge and T. Ma, "More Algorithms for Provable Dictionary Learning", arXiv:1401.0579v1

[16] J. Mairal, F. Bach, J. Ponce, G. Sapiro, and A. Zisserman, "Supervised dictionary learning", NIPS, 2008.

[17] M. Yang, L. Zhang, X. Feng, and D. Zhang, "Fisher discrimination dictionary learning for sparse representation", ICCV, 2011.

[18] Y. Sun, Q. Liu, J. Tang and D. Tao, "Learning Discriminative Dictionary for Group Sparse Representation," IEEE Transactions on Image Processing, Vol. 23 (9), pp. 3816-3828, Sept. 2014.

[19] S. Yadav, M. Singh, M. Vatsa, R. Singh and A. Majumdar, "Low Rank Group Sparse Representation Based Classifier for Pose Variation", IEEE ICIP 2016.

[20] http://rodrigob.github.io/are_we_there_yet/build/classification_datasets_results.html

[21] E. Eidinger, R. Enbar and T. Hassner. "Age and gender estimation of unfiltered faces", IEEE Transactions on Information Forensics and Security, Vol. 9 (12), pp. 2170-2179, 2014.

[22] R. Rothe, R. Timofte and L. V. Gool. "Deep expectation of real and apparent age from a single image without facial landmarks", International Journal of Computer Vision, pp. 1-14, 2016.

[23] G. Levi and T. Hassner. "Age and gender classification using convolutional neural networks". IEEE CVPR Workshop on Analysis and Modeling of Faces and Gestures, 34-42